\documentclass[11pt]{article}

\usepackage{graphicx}
\usepackage{hyperref} 

\begin{document}
\date{24 October 2021}

\title{Evolving Open Complexity}

\author{
\href{http://www.cs.ucl.ac.uk/staff/W.Langdon/}
{W. B. Langdon}
}

\maketitle

\pagestyle{plain}
\thispagestyle{empty}

\begin{abstract}
\noindent
Information theoretic analysis of large evolved programs
produced by running genetic programming
for up to a million generations has shown
even functions as smooth and well behaved as floating point
addition and multiplication
loose entropy and consequently are robust and
fail to propagate disruption to their outputs.
This means,
while dependent upon fitness tests,
many genetic changes deep within trees are silent.
For evolution to proceed at reasonable rate
it must be possible to measure the impact of most code changes,
yet in large trees most crossover sites are distant from the root node.
We suggest to evolve very large very complex programs,
it will be necessary to adopt an open architecture
where most mutation site are within 10--100 levels
of the organism's environment.
\end{abstract}

\section{Background}

Recently we have been investigating the long term evolution
of genetic programming.
Firstly, using Poli's submachine code GP
\cite{poli:1999:aigp3,poli:2000:22par},
evolving large binary Boolean trees
\cite{Langdon:2017:GECCO}
and more recently exploiting SIMD Intel AVX and multi-core
parallelism to evolve floating point GP
\cite{langdon:2019:gpquick,%
Langdon:2019:alife,%
Langdon:2021:GPTP,%
langdon:GPEM:gpconv}.
Running for up to a million generations without size limits
has generated,
at more than a billion nodes,
the biggest programs yet evolved
and forced the development
\cite{langdon:RN2001,langdon:2020:small_mem_ga_2pages}
of,
at the equivalent of more than a trillion GP operations per second,
the fastest GP system.
It has also prompted information theoretic analysis of programs
\cite{Langdon:2021:LAHS}.
(Of course information theory has long been used with
evolutionary computing, e.g.~\cite{Deschaine:2011:JMMS}.)

One immediately applicable result
has been the realisation that in deep GP trees most
changes have no impact on fitness and once this has been proved, 
for a given child,
its fitness evaluation can be cut short and
fitness simply copied from the parent.
This can lead to enormous speed ups
\cite{langdon:2021:EuroGP}.

We have also considered traditional imperative (human written)
programs
and shown these too are much more robust than is often assumed
\cite{langdon:2015:csdc,langdon:2016:dagstuhl,langdon:2017:LHS}.
Indeed we suggest that information theory provides 
a unified view of the difficulty of testing software
\cite{langdon:2021:ieeeblog,Clark:2020:facetav,Petke:2021:FSE-IVR}.

The question of why fitness is so often exactly inherited 
\cite{Petke:2019:GI7,Petke:2020:APR,Pimenta:2020:evoCOP}
despite brutal genetic change
is answered by the realisation that without side effect
the disruption caused by the mutation must be propagated
up the tree through a long chain of irreversible operations
to the root node.
Each function in the chain can loose entropy.
In many cases 
deeply nested functions progressively loose information about the perturbation
as the disruption fails to propagate to the program's output.
Thus the mutation becomes invisible to fitness testing
and its utility cannot be measured.
Without fitness selective pressure, 
evolution degenerates into an undirected random walk.

In bloated structures information loss leads,
from an evolutionary point of view,
to extremely high resilience, robustness and so stasis.
From the engineering point of view this is problematic,
as then almost all genetic changes have no impact
and evolutionary progress slows to a dawdle.

Since all digital computing is irreversible 
\cite{langdon:2007:NC}
it inherently losses
information and so without short cuts,
must lead to failed disruption propagation (FDP)~\cite{Petke:2021:FSE-IVR}.
We suggest in order to evolve large complex systems
it must be possible to measure the impact of genetic changes,
therefore we must control FDP
and suggest in the next section that
to evolve large systems,
they be composed of many programs of limited nesting depth
and structured to allow rapid communication of both inputs and
outputs to the (fitness determining) environment.

\section{Open Complex System}

Can we evolve systems more like cell interiors,
with high surface area membranes
composed of very many small adjacent programs
each of limited depth
placed side by side.
The membranes forming an open structure
with many gaps between them.
The gaps themselves supporting rapid communication
(which might be implemented using global memory
or enhanced communication links or buses)
with no, or little, processing ability
and consequently little information loss.

\begin{figure}
\centerline{\includegraphics[height=0.6\textheight]{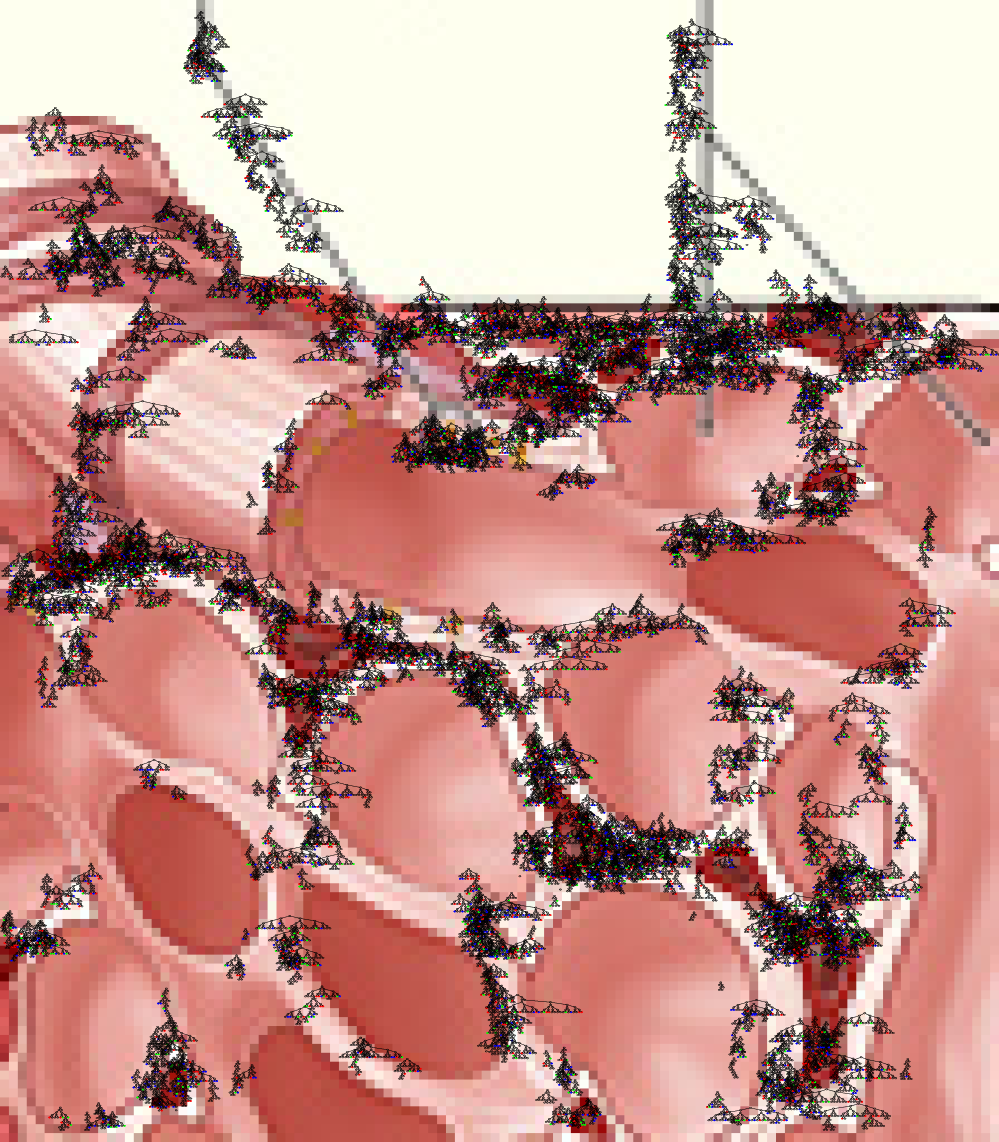}}
\caption{
Lung like open complex evolving system
composed of 1300 individual GP programs or functions. 
These compute element are placed 
side-by-side to form an open structure.
The gaps promote short cut side effects between functions'
input and outputs and the environment.
\label{fig:lung_like}
}
\end{figure}

Figure~\ref{fig:lung_like}
shows 1300 programs arranged in an open structure
(based on fit Sextic polynomial~\cite{koza:book} trees with 
average height 
9.22,
quartile range 7--11).

Rich Lenski,
in his long-term evolution experiment (LTEE)
\cite{Lenski:2015:PRoySocB,Good:2017:nature}
has demonstrated that Nature can continue
to innovate in a static environment for more than 75\,000
generations.
We have shown GP can do similarly for at least 100\,000 generations.
However when evolving deep structures, progress slows dramatically
and therefore we feel monolithic deep structures
will not be sufficient to automatically evolve complex systems.
Instead an open structure like Figure~\ref{fig:lung_like}
may be needed.

\bibliographystyle{abbrvurl1}
\bibliography{gp-bibliography,references}

\begin{thebibliography}{10}

\bibitem{Clark:2020:facetav}
D.~Clark, W.~B. Langdon, and J.~Petke.
\newblock \href
  {https://fbresearchevents.bevylabs.com/events/details/facebook-tav-symposium-division-facebook-testing-and-verification-symposium-presents-dress-rehearsal-facebook-tav-symposium-2020/}
  {Software robustness: A survey, a theory, and some prospects}.
\newblock Presented at Facebook Testing and Verification Symposium 2020, 1-3
  December 2020.

\bibitem{Deschaine:2011:JMMS}
L.~M. Deschaine, P.~Nordin, and J.~D. Pinter.
\newblock \href {http://www.immsp.kiev.ua/publications/eng/2011_2/} {A
  computational geometric/information theoretic method to invert physics-based
  {MEC} models attributes for {MEC} discrimination}.
\newblock {\em Journal of Mathematical Machines and Systems}, (2):50--61, 2011.

\bibitem{Good:2017:nature}
B.~H. Good, M.~J. McDonald, J.~E. Barrick, R.~E. Lenski, and M.~M. Desai.
\newblock \href {http://dx.doi.org/doi:10.1038/nature24287} {The dynamics of
  molecular evolution over 60,000 generations}.
\newblock {\em Nature}, 551:45--50, 18 October 2017.

\bibitem{koza:book}
J.~R. Koza.
\newblock \href {http://mitpress.mit.edu/books/genetic-programming} {{\em
  Genetic Programming: On the Programming of Computers by Means of Natural
  Selection}}.
\newblock MIT Press, Cambridge, MA, USA, 1992.

\bibitem{langdon:GPEM:gpconv}
W.~B. Langdon.
\newblock \href {http://dx.doi.org/doi:10.1007/s10710-021-09405-9} {Genetic
  programming convergence}.
\newblock {\em Genetic Programming and Evolvable Machines}.

\bibitem{langdon:2016:dagstuhl}
W.~B. Langdon.
\newblock \href {http://dx.doi.org/doi:10.4230/DagRep.5.10.89} {{\em Software
  is not fragile}}, volume~5, chapter 4.7, pages 100--101.
\newblock Schloss Dagstuhl--Leibniz-Zentrum fuer Informatik, Dagstuhl, Germany,
  2016.

\bibitem{Langdon:2017:GECCO}
W.~B. Langdon.
\newblock \href {http://dx.doi.org/doi:10.1145/3067695.3075965} {Long-term
  evolution of genetic programming populations}.
\newblock In {\em Proceedings of the Genetic and Evolutionary Computation
  Conference Companion}, GECCO '17, pages 235--236, Berlin, 15-19 July 2017.
  ACM.

\bibitem{langdon:2017:LHS}
W.~B. Langdon.
\newblock Long-term stability of genetic programming landscapes.
\newblock In N.~Veerapen, F.~Daolio, A.~Liefooghe, S.~Verel, and G.~Ochoa,
  editors, {\em GECCO 2017 Workshop on Landscape-Aware Heuristic Search},
  Berlin, 2017.
\newblock No paper, presentation only.

\bibitem{langdon:2019:gpquick}
W.~B. Langdon.
\newblock \href {http://dx.doi.org/doi:10.1145/3319619.3326770} {Parallel
  {GPQUICK}}.
\newblock In C.~Doerr, editor, {\em GECCO '19: Proceedings of the Genetic and
  Evolutionary Computation Conference Companion}, pages 63--64, Prague, Czech
  Republic, July 13-17 2019. ACM.

\bibitem{langdon:RN2001}
W.~B. Langdon.
\newblock \href {https://arxiv.org/abs/2001.04505} {Fast generation of big
  random binary trees}.
\newblock Technical Report RN/20/01, Computer Science, University College,
  London, Gower Street, London, UK, 13 Jan. 2020.

\bibitem{langdon:2020:small_mem_ga_2pages}
W.~B. Langdon.
\newblock \href {http://dx.doi.org/doi:10.1145/3430913.3430914} {Multi-threaded
  memory efficient crossover in {C++} for generational genetic programming}.
\newblock {\em SIGEVOLution newsletter of the ACM Special Interest Group on
  Genetic and Evolutionary Computation}, 13(3):2--4, Oct. 2020.

\bibitem{Langdon:2021:GPTP}
W.~B. Langdon.
\newblock \href
  {http://www.cs.ucl.ac.uk/staff/W.Langdon/ftp/papers/Langdon_2021_GPTP.pdf}
  {Fitness first}.
\newblock In W.~Banzhaf, L.~Trujillo, S.~Winkler, and B.~Worzel, editors, {\em
  Genetic Programming Theory and Practice XVIII}, East Lansing, MI, USA, 19-21
  May 2021. Springer.
\newblock Forthcoming.

\bibitem{langdon:2021:EuroGP}
W.~B. Langdon.
\newblock \href {http://dx.doi.org/doi:10.1007/978-3-030-72812-0_15}
  {Incremental evaluation in genetic programming}.
\newblock In T.~Hu, N.~Lourenco, and E.~Medvet, editors, {\em EuroGP 2021:
  Proceedings of the 24th European Conference on Genetic Programming}, volume
  12691 of {\em LNCS}, pages 229--246, Virtual Event, 7-9 Apr. 2021. Springer
  Verlag.

\bibitem{Langdon:2019:alife}
W.~B. Langdon and W.~Banzhaf.
\newblock \href {http://dx.doi.org/doi:10.1162/isal_a_00191} {Continuous
  long-term evolution of genetic programming}.
\newblock In R.~Fuechslin, editor, {\em Conference on Artificial Life (ALIFE
  2019)}, pages 388--395, Newcastle, 29 July - 2 Aug. 2019. MIT Press.

\bibitem{langdon:2015:csdc}
W.~B. Langdon and J.~Petke.
\newblock \href {http://dx.doi.org/doi:10.1007/978-3-319-45901-1_24} {Software
  is not fragile}.
\newblock In P.~Parrend, P.~Bourgine, and P.~Collet, editors, {\em Complex
  Systems Digital Campus E-conference, CS-DC'15}, Proceedings in Complexity,
  pages 203--211. Springer, Sept. 30-Oct. 1 2015.
\newblock Invited talk.

\bibitem{Langdon:2021:LAHS}
W.~B. Langdon, J.~Petke, and D.~Clark.
\newblock \href {http://dx.doi.org/doi:10.1145/3449726.3463147} {Dissipative
  polynomials}.
\newblock In N.~Veerapen, K.~Malan, A.~Liefooghe, S.~Verel, and G.~Ochoa,
  editors, {\em 5th Workshop on Landscape-Aware Heuristic Search}, GECCO 2021
  Companion, pages 1683--1691, Internet, 10-14 July 2021. ACM.

\bibitem{langdon:2021:ieeeblog}
W.~B. Langdon, J.~Petke, and D.~Clark.
\newblock \href
  {http://blog.ieeesoftware.org/2021/09/information-loss-leads-to-robustness-w.html}
  {Information loss leads to robustness}.
\newblock IEEE Software Blog, 12 September 2021.

\bibitem{langdon:2007:NC}
W.~B. Langdon and R.~Poli.
\newblock \href {http://dx.doi.org/doi:10.1007/s11047-007-9044-x} {Mapping
  non-conventional extensions of genetic programming}.
\newblock {\em Natural Computing}, 7:21--43, Mar. 2008.
\newblock Invited contribution to special issue on Unconventional computing.

\bibitem{Lenski:2015:PRoySocB}
R.~E. Lenski et~al.
\newblock \href {http://dx.doi.org/doi:10.1098/rspb.2015.2292} {Sustained
  fitness gains and variability in fitness trajectories in the long-term
  evolution experiment with {Escherichia} coli}.
\newblock {\em Proceedings of the Royal Society B}, 282(1821), 22 December
  2015.

\bibitem{Petke:2019:GI7}
J.~Petke, B.~Alexander, E.~T. Barr, A.~E.~I. Brownlee, M.~Wagner, and D.~R.
  White.
\newblock \href {http://dx.doi.org/doi:10.1145/3319619.3326870} {A survey of
  genetic improvement search spaces}.
\newblock In B.~Alexander, S.~O. Haraldsson, M.~Wagner, and J.~R. Woodward,
  editors, {\em 7th edition of GI @ GECCO 2019}, pages 1715--1721, Prague,
  Czech Republic, July 13-17 2019. ACM.

\bibitem{Petke:2020:APR}
J.~Petke and A.~Blot.
\newblock \href {http://dx.doi.org/doi:10.1145/3387940.3392180} {Refining
  fitness functions in test-based program repair}.
\newblock In S.~H. Tan, S.~Mechtaev, M.~Monperrus, and M.~Prasad, editors, {\em
  The First International Workshop on Automated Program Repair (APR@ICSE
  2020)}, pages 13--14, internet, 2020. Association for Computing Machinery.

\bibitem{Petke:2021:FSE-IVR}
J.~Petke, D.~Clark, and W.~B. Langdon.
\newblock \href {http://dx.doi.org/doi:10.1145/3468264.3473133} {Software
  robustness: A survey, a theory, and some prospects}.
\newblock In P.~Avgeriou and D.~Zhang, editors, {\em ESEC/FSE 2021, Ideas,
  Visions and Reflections}, pages 1475--1478, Athens, Greece, 23-28 Aug. 2021.
  ACM.

\bibitem{Pimenta:2020:evoCOP}
C.~G. Pimenta, A.~G.~C. {de Sa}, G.~Ochoa, and G.~L. Pappa.
\newblock \href {http://dx.doi.org/doi:10.1007/978-3-030-43680-3_8} {Fitness
  landscape analysis of automated machine learning search spaces}.
\newblock In L.~Paquete and C.~Zarges, editors, {\em European Conference on
  Evolutionary Computation in Combinatorial Optimization (EvoCOP 2020)}, volume
  12102 of {\em Lecture Notes in Computer Science}, pages 114--130, Seville,
  Spain, 15-17 Apr. 2020. Springer.

\bibitem{poli:1999:aigp3}
R.~Poli and W.~B. Langdon.
\newblock \href {http://www.cs.ucl.ac.uk/staff/W.Langdon/aigp3/ch13.pdf}
  {Sub-machine-code genetic programming}.
\newblock In L.~Spector, W.~B. Langdon, U.-M. O'Reilly, and P.~J. Angeline,
  editors, {\em Advances in Genetic Programming 3}, chapter~13, pages 301--323.
  MIT Press, Cambridge, MA, USA, June 1999.

\bibitem{poli:2000:22par}
R.~Poli and J.~Page.
\newblock \href {http://dx.doi.org/doi:10.1023/A:1010068314282} {Solving
  high-order {Boolean} parity problems with smooth uniform crossover,
  sub-machine code {GP} and demes}.
\newblock {\em Genetic Programming and Evolvable Machines}, 1(1/2):37--56, Apr.
  2000.

\end{thebibliography}

\end{document}